\documentclass{article}

\usepackage[numbers]{natbib}

\usepackage[preprint]{neurips_data_2021}

\usepackage[utf8]{inputenc} 
\usepackage[T1]{fontenc}    
\usepackage{hyperref}       
\usepackage{url}            
\usepackage{booktabs}       
\usepackage{amsfonts}       
\usepackage{nicefrac}       
\usepackage{microtype}      
\usepackage{xcolor}         
\usepackage{graphics}
\usepackage{subfig}
\usepackage{graphicx}
\usepackage{amsmath}
\usepackage{tabularx}
\usepackage{array}
\usepackage{multirow}
\usepackage{CJKutf8}
\usepackage{bbold}
\usepackage{enumitem}
\usepackage{amssymb}
\usepackage{pifont}
\newcommand{\cmark}{\ding{51}}%
\newcommand{\xmark}{\ding{55}}%

\usepackage{tikz}
\usepackage{adjustbox}
\usepackage{bold-extra}
\usepackage{CJKutf8}

\newcommand{\dataset}{\bbfamily BiToD}

\title{{\dataset}: A Bilingual Multi-Domain Dataset For Task-Oriented Dialogue Modeling}

\author{%
  Zhaojiang Lin$^1$\thanks{\quad Equal contribution}$^*$, Andrea Madotto$^1$$^*$, Genta Indra Winata$^1$, Peng Xu$^1$,\\ \textbf{Feijun Jiang$^2$, Yuxiang Hu$^2$, Chen Shi$^2$, Pascale Fung$^1$} \\
  $^1$Center for Artificial Intelligence Research (CAiRE)\\
  $^1$The Hong Kong University of Science and Technology\\
  $^2$Alibaba Group \\
  \texttt{\{zlinao, amadotto, giwinata, pxuab\}@connect.ust.hk} \\
}

\begin{document}

\maketitle

\begin{abstract}

Task-oriented dialogue (ToD) benchmarks provide an important avenue to measure progress and develop better conversational agents. However, existing datasets for end-to-end ToD modeling are limited to a single language, hindering the development of robust end-to-end ToD systems for multilingual countries and regions. Here we introduce {\dataset}\footnote{Data and code are available in \url{https://github.com/HLTCHKUST/BiToD}.}, the first bilingual multi-domain dataset for end-to-end task-oriented dialogue modeling. {\dataset} contains over 7k multi-domain dialogues (144k utterances) with a large and realistic bilingual knowledge base. It serves as an effective benchmark for evaluating bilingual ToD systems and cross-lingual transfer learning approaches. We provide state-of-the-art baselines under three evaluation settings (monolingual, bilingual, and cross-lingual). The analysis of our baselines in different settings highlights 1) the effectiveness of training a bilingual ToD system compared to two independent monolingual ToD systems, and 2) the potential of leveraging a bilingual knowledge base and cross-lingual transfer learning to improve the system performance under low resource conditions.

\end{abstract}

\section{Introduction}
Task-oriented dialogue (ToD) systems are designed to assist humans in performing daily activities, such as ticket booking, travel planning, and online shopping. These systems are the core modules of virtual assistants (e.g., Apple Siri and Amazon Alexa), and they provide natural language interfaces for online services~\cite{rastogi2020towards}. Recently, there has been growing interest in developing deep learning-based end-to-end ToD systems \cite{bordes2016learning,wen2017network,eric2017key,qin2019entity,qin2020dynamic,banerjee2019graph,neelakantan2019neural,eric2017copy,madotto2018mem2seq,reddy2019multi,wu2019global,hosseini2020simple,peng2020soloist,lin2020mintl,byrne2020tickettalk} because they can handle complex dialogue patterns with minimal hand-crafted rules. To advance the existing state-of-the-art, large-scale datasets~\cite{budzianowski2018multiwoz,rastogi2020towards,byrne2020tickettalk} have been proposed for training and evaluating such data-driven systems.

However, existing datasets for end-to-end ToD modelling are limited to a single language, such as English~\cite{budzianowski2018multiwoz,mosig2020star}, or Chinese~\cite{zhu2020crosswoz,quan2020risawoz}. The absence of bilingual or multilingual datasets not only limits the research on cross-lingual transfer learning~\cite{razumovskaia2021crossing} but also hinders the development of robust end-to-end ToD systems for multilingual countries and regions. 

To tackle the challenge mentioned above, we introduce {\dataset}, a bilingual multi-domain dataset for task-oriented dialogue modelling. {\dataset} has 7,232 bilingual dialogues (in English and Chinese), spanning seven services within five domains, where each dialogue is annotated with dialogue states, speech-acts, and service API calls.  Therefore, {\dataset} can be used for building both end-to-end ToD systems and dialogue sub-modules (e.g., Dialogue State Tracking). 
We propose three evaluation settings: 1) \textbf{\textit{monolingual}}, in which the models are trained and tested on either English or Chinese data, 2) \textbf{\textit{bilingual}}, where the models are trained with bilingual data and tested with English and Chinese dialogues simultaneously, and 3) \textbf{\textit{cross-lingual}}, where the models are first trained with the source language and then tested in a few-shot setting in the target language.

\begin{figure}%
\begin{center}
\begin{adjustbox}{width=0.48\textwidth}
\begin{tikzpicture}
\definecolor{chatcolor1}{HTML}{E2F0FF}
\definecolor{chatcolor2}{HTML}{FDF0FF}
\definecolor{chatcolor3}{HTML}{FFC9CF}
\definecolor{chatcolor4}{HTML}{E6E6E6}
{\small
\fontfamily{cmss}\selectfont
\node[align=left, text width=6cm, fill=chatcolor2, rounded corners=1mm, anchor=north west] at (2,0) {\includegraphics[height=0.29cm]{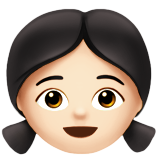}: Hi, can you help me find a place to eat?};
\node[align=left, text width=6cm,fill=chatcolor1, rounded corners=1mm, anchor=north west] at (0,-.97) {\includegraphics[height=0.29cm]{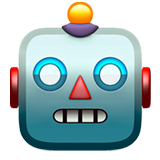}: Sure! How much do you want to spend and how high of a rating would you prefer?};
\node[align=left, text width=6cm, fill=chatcolor2, rounded corners=1mm, anchor=north west] at (2,-1.99) {\includegraphics[height=0.29cm]{img/emoji/girl.png}: I'd like to eat at an expensive restaurant rated at least 9.};
\node[align=left, text width=6cm,fill=chatcolor1, rounded corners=1mm, anchor=north west] at (0,-2.94) {\includegraphics[height=0.29cm]{img/emoji/robot.png}: Got it. What kind of food do you want?};
\node[align=left, text width=6cm, fill=chatcolor2, rounded corners=1mm, anchor=north west] at (2,-3.87) {\includegraphics[height=0.29cm]{img/emoji/girl.png}: Any type of food is fine, but I want a place with Vegan Options.};
\node[align=left, text width=6.5cm,fill=chatcolor3, rounded corners=1mm, anchor=north west] at (0,-4.87) {\includegraphics[height=0.29cm]{img/emoji/robot.png}\includegraphics[height=0.29cm]{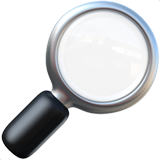}: \texttt{RESTAURANTS\_SEARCH(dietary=Vegan,\\ \quad\quad \ rating$\geq$9, price=expensive)}};
\node[align=left, text width=6.5cm,fill=chatcolor4, rounded corners=1mm, anchor=north west] at (0,-5.85) { \texttt{\textbf{name}: Sushi Tsubomi\\
\textbf{location}: \framebox[\width]{\textit{Wan Chai}}\\
\textbf{cuisine}: Japanese\\
\textbf{price}: expensive\\
\textbf{rating}: 10\\
\textbf{dietary}: Vegan\\
\textbf{available options}: 51\\}};
\node[align=left, text width=6cm,fill=chatcolor1, rounded corners=1mm, anchor=north west] at (0,-8.49) {\includegraphics[height=0.29cm]{img/emoji/robot.png}: Okay. Out of 51 restaurants, I recommend you go to Sushi Tsubomi. They are a Japanese restaurant rated at a 10.};
\node[align=center, text width=1cm, rounded corners=1mm, anchor=north west] at (2,-9.8) { {\Large$\cdots$} };
\node[align=left, text width=6cm, fill=chatcolor2, rounded corners=1mm, anchor=north west] at (2,-10.25) {\includegraphics[height=0.29cm]{img/emoji/girl.png}: I'd like to visit an attraction nearby as well, rated at an 8 minimum.};
\node[align=left, text width=6cm,fill=chatcolor1, rounded corners=1mm, anchor=north west] at (0,-11.25) {\includegraphics[height=0.29cm]{img/emoji/robot.png}: Sure, I can help with that. What type of attraction would you like?};
\node[align=left, text width=6cm, fill=chatcolor2, rounded corners=1mm, anchor=north west] at (2,-12.19) {\includegraphics[height=0.29cm]{img/emoji/girl.png}: I'd like to visit some famous landmarks.};
\node[align=left, text width=6cm,fill=chatcolor3, rounded corners=1mm, anchor=north west] at (0,-12.8) {\includegraphics[height=0.29cm]{img/emoji/robot.png}\includegraphics[height=0.29cm]{img/emoji/glass.png}: \texttt{ATTRACTIONS\_SEARCH(rating$\geq$8, type=Landmarks, location=\framebox[\width]{\textit{Wan Chai}})}};
\node[align=left, text width=6cm,fill=chatcolor4, rounded corners=1mm, anchor=north west] at (0,-13.96) { \texttt{\textbf{name}: Hong Kong Convention $\cdots$\\
\textbf{location}: Wan Chai\\
\textbf{rating}: 8\\
\textbf{type}: Sights \& Landmarks\\
\textbf{address}: 1 Expo Dr., Wanchai\\
\textbf{phone}: 2582 8888\\
\textbf{available options}: 5\\}};
\node[align=center, text width=1cm, rounded corners=1mm, anchor=north west] at (2,-16.47) {};
}

\end{tikzpicture}
\end{adjustbox}
\quad
\begin{adjustbox}{width=0.48\textwidth}
\begin{tikzpicture}
\definecolor{chatcolor1}{HTML}{E2F0FF}
\definecolor{chatcolor2}{HTML}{FDF0FF}
\definecolor{chatcolor3}{HTML}{FFC9CF}
\definecolor{chatcolor4}{HTML}{E6E6E6}
{\small
\fontfamily{cmss}\selectfont
\begin{CJK*}{UTF8}{gkai}
\node[align=left, text width=6cm, fill=chatcolor2, rounded corners=1mm, anchor=north west] at (2,0) {\includegraphics[height=0.29cm]{img/emoji/girl.png}: 你好，我想订一家餐厅吃饭，我不想吃法餐。};
\node[align=left, text width=6cm,fill=chatcolor1, rounded corners=1mm, anchor=north west] at (0,-0.97) {\includegraphics[height=0.29cm]{img/emoji/robot.png}: 好的，请问您是在哪里找餐厅？要什么价位的呢？};
\node[align=left, text width=6cm, fill=chatcolor2, rounded corners=1mm, anchor=north west] at (2,-1.96) {\includegraphics[height=0.29cm]{img/emoji/girl.png}: 位置和价位都不重要。但需要在9分以上。};
\node[align=left, text width=6cm,fill=chatcolor3, rounded corners=1mm, anchor=north west] at (0,-2.87) {\includegraphics[height=0.29cm]{img/emoji/robot.png}\includegraphics[height=0.29cm]{img/emoji/glass.png}: \texttt{餐馆查询(评分$\geq$9, 菜品≠法餐)}};
\node[align=left, text width=6cm,fill=chatcolor4, rounded corners=1mm, anchor=north west] at (0,-3.47) { \texttt{名字: ChocoDuck Bistro\\
位置: 新界\\
菜品: 意餐\\
价格范围: 适中\\
评分 :10\\
饮食限制: 适合素食主义者\\
地址: 3 Kin San Lane, Cheung Chau\\
电话: 9735 0049\\
可用选项: 463\\}};
\node[align=left, text width=6cm,fill=chatcolor1, rounded corners=1mm, anchor=north west] at (0,-6.82) {\includegraphics[height=0.29cm]{img/emoji/robot.png}: 有463家不错的餐厅，其中有一家评分为10分，名字叫ChocoDuck Bistro，他主营意餐。};
\node[align=left, text width=6cm, fill=chatcolor2, rounded corners=1mm, anchor=north west] at (2,-8.1) {\includegraphics[height=0.29cm]{img/emoji/girl.png}: 还有别的吗};
\node[align=left, text width=6cm,fill=chatcolor3, rounded corners=1mm, anchor=north west] at (0,-8.78) {\includegraphics[height=0.29cm]{img/emoji/robot.png}\includegraphics[height=0.29cm]{img/emoji/glass.png}: \texttt{餐馆查询(菜品$\neq$法餐, 评分$\geq$9, 名字$\neq$ChocoDuck Bistro)} };
\node[align=left, text width=6cm,fill=chatcolor4, rounded corners=1mm, anchor=north west] at (0,-9.77) { \texttt{名字: Hooked\\
位置: 中区\\
菜品: 海鲜\\
价格范围: 便宜\\
评分: 10\\
饮食限制: null\\
地址: 半山坚道86-88号\\
电话: 2915 1118\\
可用选项: 462\\}};
\node[align=left, text width=6cm,fill=chatcolor1, rounded corners=1mm, anchor=north west] at (0,-13.2) {\includegraphics[height=0.29cm]{img/emoji/robot.png}: 还有462家，推荐您可以试试Hooked，它以海鲜为主，评分是10分};
\node[align=left, text width=6cm, fill=chatcolor2, rounded corners=1mm, anchor=north west] at (2,-14.15) {\includegraphics[height=0.29cm]{img/emoji/girl.png}: 好，帮我订一下，星期二下午5:50的。};
\node[align=left, text width=6cm,fill=chatcolor1, rounded corners=1mm, anchor=north west] at (0,-14.73) {\includegraphics[height=0.29cm]{img/emoji/robot.png}: 好的，请问您几位用餐？怎么称呼您？};
\node[align=left, text width=6cm, fill=chatcolor2, rounded corners=1mm, anchor=north west] at (2,-15.3) {\includegraphics[height=0.29cm]{img/emoji/girl.png}: 马连良，5位。};
\node[align=left, text width=6cm,fill=chatcolor1, rounded corners=1mm, anchor=north west] at (0,-15.87) {\includegraphics[height=0.29cm]{img/emoji/robot.png}: 好的，跟您确认一下，为您预订星期二下午5:50，Hooked餐厅5位用餐。};

\end{CJK*}
}
\end{tikzpicture}
\end{adjustbox}
\end{center}
    \caption{Dialogue examples in English (on the left) and Chinese (on the right) from the {\dataset} dataset (best viewed in colour). The user (in pink) asks for restaurant and attraction information. At the same time, the system (in blue) responds to the user utterance by calling APIs (in red) when needed and by incorporating the API results (in grey). }%
    \label{fig:example}%
\end{figure}

The contribution of this work is three-fold. 1) We propose the first bilingual dataset ({\dataset}) with a total of 7, 232 dialogues for \textit{\textbf{end-to-end}} ToD modeling. {\dataset} serves as an effective benchmark for evaluating bilingual ToD systems and cross-lingual transfer learning approaches. 2) We provide novel baselines under the three evaluation settings, i.e., monolingual, bilingual, and cross-lingual. 3) We show the effectiveness of training a bilingual ToD system compared to two independent monolingual ToD systems as well as the potential of leveraging a bilingual knowledge base and cross-lingual transfer learning to improve the system performance under low resource condition.

    
 

The paper is organized as follows: We next describe the {\dataset} data collection methods in Section 2. We then describe our proposed tasks in section 3. Section 4 introducew our baselines, and we finally present and discuss results in Section 5.

\section{{\dataset} Dataset}
{\dataset} is designed to develop virtual assistants in multilingual cities, regions, or countries (e.g., Singapore, Hong Kong, India, Switzerland, etc.). For the {\dataset} data collection, we chose Hong Kong since it is home to plenty of attractions, restaurants and more, and is one of the most visited cities globally, especially by English and Chinese speakers. This section describes the knowledge base construction and provides detailed descriptions of the dialogue collection. 


\subsection{Knowledge Base Collection}
We collect publicly available Hong Kong tourism information from the Web, to create a knowledge base that includes 98 metro stations, 305 attractions, 699 hotels, and 1,218 restaurants. For the weather domain, we synthetically generate the weather information on different dates. Then, we implement seven service APIs (\texttt{Restaurant\_Searching}, \texttt{Restaurant\_Booking}, \texttt{Hotel\_Searching}, \texttt{Hotel\_Booking}, \texttt{Attraction\_Searching}, \texttt{MTR\_info}, \texttt{Weather\_info}) to query our knowledge base. The knowledge base statistics are shown in Table \ref{tab:kb}.  Although we aim to collect a fully parallel knowledge base, we observe that some items do not include bilingual information. For example, several traditional Cantonese restaurants do not have English names, and similarly, some restaurants do not provide addresses in Chinese. This lack of parallel information reflects the real-world challenges that databases are often incomplete and noisy. 

\begin{figure}[t]
    \centering
    \includegraphics[width=\linewidth]{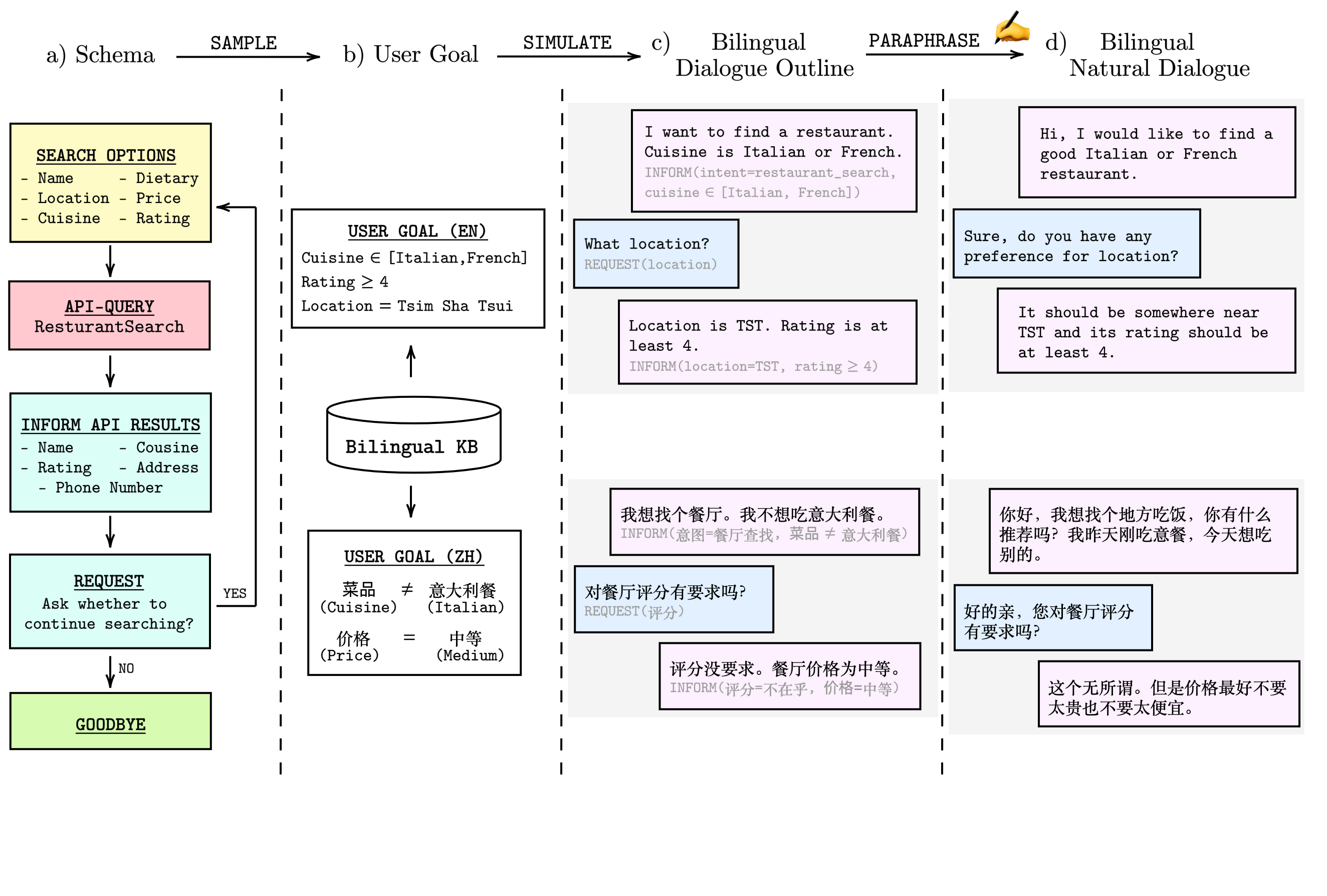}
    \caption{Illustration of the bilingual dialogues collection pipeline: a) Design a schema for each service API; b) Sample user goals from the bilingual knowledge base (KB) according to schema; c) Based on one of the user goals, the dialogue simulator generates the dialogue outlines while interacting with the APIs; d) Convert the dialogue outlines to natural conversations via crowdsourcing. Note that English and Chinese user goals are sampled independently.}
    \label{fig:data_collection}
\end{figure}

\subsection{Dialogue Data Collection}
The dialogues are collected through a four-phase pipeline, as shown in Figure \ref{fig:data_collection}. We first design a schema, as a flowchart, for each service API, to specify the possible API queries and expected system actions after the API call. Then, user goals are sampled from the knowledge base according to the pre-defined schemas. Based on the user goals, the dialogue simulator interacts with the APIs to generate dialogue outlines. 
Finally, the dialogue outlines are converted into natural conversations through crowdsourcing. Our data collection methodology extends the Machine-to-Machine (M2M)  approaches~\cite{shah2018building,rastogi2020towards} to bilingual settings to minimize the annotation overhead (time and cost).

\begin{table}[t]
\centering
\begin{tabular}{@{}lcccccccccc@{}}
\toprule
\textbf{Domain}          & \multicolumn{2}{c}{\textbf{Attraction}} & \multicolumn{2}{c}{\textbf{Hotel}} & \multicolumn{2}{c}{\textbf{Restaurant}} & \multicolumn{2}{c}{\textbf{Weather}} & \multicolumn{2}{c}{\textbf{Metro}} \\ \midrule
Language & EN & ZH & EN & ZH & EN & ZH & EN & ZH & EN & ZH \\ \midrule
\# Slots        & 7              & 7              & 9            & 9           & 13             & 13             & 5             & 5            & 5          & 5          \\
\# Entities     & 1,079           & 1,118           & 2,642         & 2,652        & 5,489           & 5,035           & 77            & 77           & 161        & 161        \\ \bottomrule
\end{tabular}
\caption{Bilingual knowledge base statistics.}
\label{tab:kb}
\end{table}

\paragraph{Schemas and APIs.}
The dialogue schema shown as a flowchart (\texttt{Restaurant\_Searching}) in Figure \ref{fig:data_collection}.a specifies the input and output options of the API and the desired system behaviours. To elaborate, the user searches a restaurant by name, location, cuisine, etc. Then the system calls the API and informs the user of the restaurant name and other requested information. If the user is not satisfied with the search results, the system continues searching and provides other options. To ensure the provided services are realistic, we impose a few restrictions, as in \cite{rastogi2020towards}. Firstly, each API has a list of required slots, and the system is not allowed to hit the API without specifying values for these slots. For example, the system needs to obtain departure and destination locations before calling the metro-info API. Secondly, the system must confirm the booking information with the user before making any reservations (e.g., restaurant booking).

\begin{figure}[t]
    \centering
    \includegraphics[width=\linewidth]{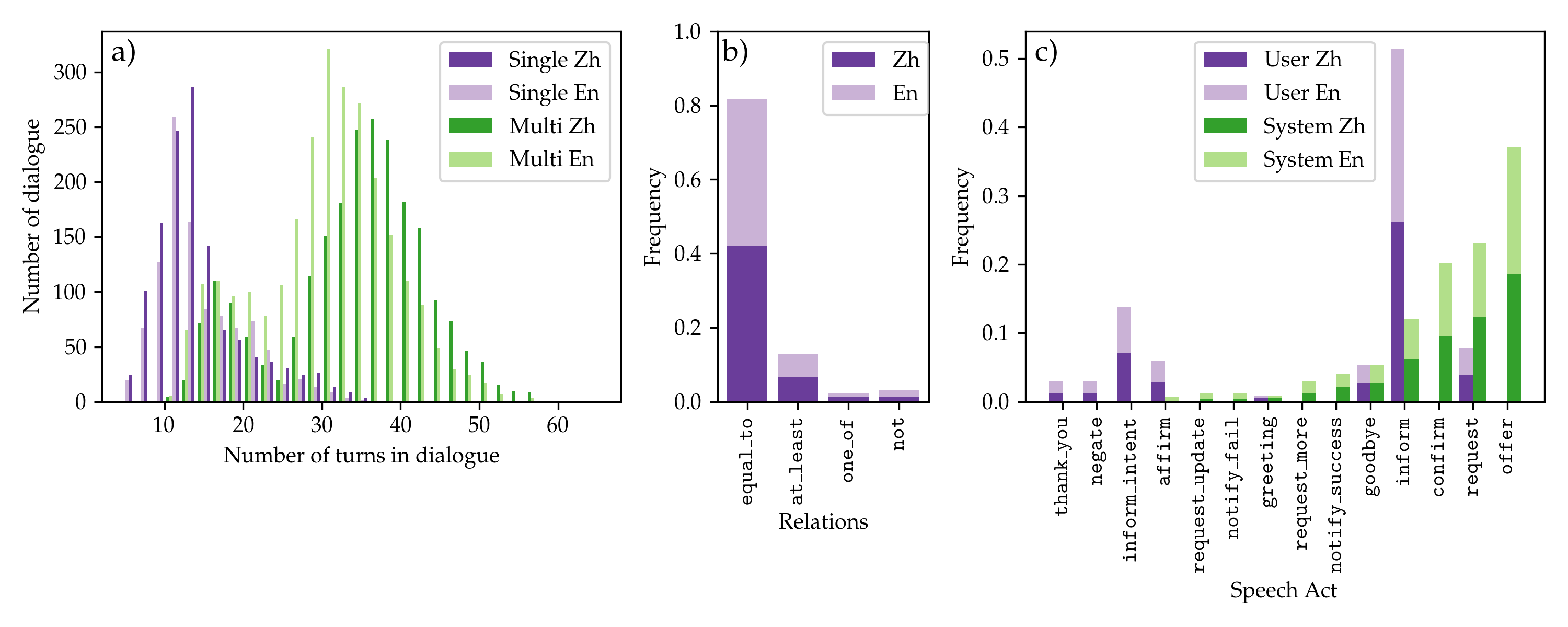}
    \caption{Data statistic of {\dataset}: a) dialogue distribution of lengths of single and multi-domain dialogues, b) distribution of different relation types, and c) distribution of speech acts of users and systems.}
    \label{fig:histo}
\end{figure}

\paragraph{User Goals.}
A user goal consists of a list of intents and a set of constraints under each intent. Figure \ref{fig:data_collection}.b shows a single domain (intent) example where the user's intent is \textit{Restaurant\_Search}. A constraint is defined with a triple (slot, relation, value) (e.g., \texttt{(Rating, at\_least, 4)}).  Different from previous work, which defined user constraints as slot-value pairs, we impose slot-value relations~\cite{mosig2020star} (listed in Figure~\ref{fig:histo}.b) to promote more diverse user goals. To generate a user goal, we first sample a list of intents. We randomly sample a set of slot-relation-value combinations from the bilingual knowledge base for each intent, which includes non-existent combinations to create unsatisfiable user requests. In multi-domain scenarios, we set a certain probability to share the same values for some of the cross-domain slots (e.g., date and location) to make the transition among domains smooth. For example, users might want to book restaurants and hotels on the same date or take the metro from the location of the booked restaurant to their hotel. Note that the user goals for English and Chinese are sampled independently, as the real-world customer service conversations are often unparalleled.

\paragraph{Dialogue Outline Generation.}
Dialogue outlines are generated by a bilingual dialogue simulator that accepts user goals in both languages as inputs. The dialogue simulator consists of a user agent and a system agent. Both agents interact with each other using a finite set of actions specified by speech acts over a probabilistic automaton designed to capture varied dialogue trajectories \cite{rastogi2020towards}. Each speech act takes a slot or slot-relation-value triple as an argument. When the conversation starts, the user agent is assigned a goal, while the system agent is initialized with a set of requests related to the services. During the conversation, the user informs constraints according to the user goal, and the system responds to the user queries while interacting with the service APIs. For some services, the system needs to request all the required slots before querying the APIs. After the API call, the system either informs the search result or searches for other options until the user intents are fulfilled. Following \cite{rastogi2020towards}, we also augment the value entities during the dialogue outlines generation process, e.g., \textit{\textbf{Tsim Sha Tsui}} can be replaced with its abbreviation \textit{\textbf{TST}}, as shown in Figure \ref{fig:data_collection}.c. After the user goal is fulfilled by a series of user and system actions, we convert all the actions into natural language using templates. In this phase, we obtain the dialogue states annotations and speech acts automatically for both the user and system sides.

\paragraph{Dialogue Paraphrase.} The dialogue outlines are converted to natural dialogues via crowdsourcing. Figure \ref{fig:zh_interface} and \ref{fig:en_interface} show the interface for Chinese and English paraphrasing, where workers see the full dialogue and rewrite the dialogue turn by turn. Before the task, workers are asked to read the instructions, shown in Figure \ref{fig:en_instructions}. In the instructions, we specify that the paraphrased dialogue should retain the same meaning as the dialogue outline but sound like a real conversation between a user and a professional assistant. The user utterances are expected to be creative and diverse, while the system utterances are expected to be formal and correct. To ensure all the essential information is presented in the new dialogue, we highlight all the entities with bold text. In the user utterances, the highlighted entities are allowed to be paraphrased without losing their original meaning; e.g., \textit{``The restaurant should provide \textbf{Vegan Options}"} is allowed to be rewritten as \textit{``I would like to find a \textbf{vegan-friendly} restaurant"}. In contrast, all the entities in the system utterances are required to be unchanged.

\paragraph{Quality Verification.} After the dialogue paraphrasing, workers are asked to read through the new dialogue and answer the following questions, as in ~\cite{shah2018building}: 1) \textit{Does it seem like a conversation between a user that sounds like you and an assistant that sounds formal?} 2) \textit{Does it have the same meaning as the original conversation, while still making sense on its own?} The first question is for examining whether the new conversation is realistic, and the second question is for verifying whether the dialogue outline and the paraphrased dialogue are valid. Given the two answer options: 1) \textit{Yes}, 2) \textit{No, but I cannot make it better}, 97.56\% of annotators chose the first option for the first question and 98.89\% of them chose the first option for the second question.

\subsection{Dataset Statistics}
We collected 7,232 dialogues with 144,798 utterances, in which 3,689 dialogues are in English, and 3,543 dialogues are in Chinese. We split the data into 80\% training, 8\% validation, and 12\% testing, resulting in 5,787 training dialogues, 542 validation dialogues, and 902 testing dialogues. In Figure~\ref{fig:histo} we show the main data statistics of the BiToD corpus. As shown in Figure~\ref{fig:histo}.a, the lengths of the dialogues vary from 10 turns to more than 50 turns. Multi-domain dialogues, in both English and Chinese, have many more turns compared to single-domains. The most used relation in user goals is \texttt{equal\_to} (Figure~\ref{fig:histo}.b), and the most common speech-acts (Figure~\ref{fig:histo}.c) for users and systems are \texttt{inform} and \texttt{offer}, respectively. Finally, in Table~\ref{tab:ontology} in the Appendix, we list all the informable and requestable slots per domain. 


\subsection{Dataset Features}
Table \ref{tab:comparison} shows the comparison of the {\dataset} training set to previous ToD datasets. Prior work for end-to-end ToD modelling only focuses on a single language. Our {\dataset} is the first \textbf{\textit{bilingual}} ToD corpus with comparable data size. In addition to its bilingualism, {\dataset} also provides the following unique features:

\paragraph{Deterministic API.} Given an API query for recommendation services (e.g., restaurant searching and hotel searching), there is typically more than one matched item. Previous works~\cite{budzianowski2018multiwoz,zhu2020crosswoz} have randomly sampled one or two items as API results and returned them to users. However, in real-world applications, the system should recommend items according to certain criteria (e.g., user rating). 
Moreover, the randomness of the API also increases the difficulty of evaluating the models. Indeed, the evaluation metrics in \cite{budzianowski2018multiwoz,zhu2020crosswoz} rely on delexicalized response templates, which are not compatible with knowledge-grounded generation approaches~\cite{madotto2018mem2seq,wu2019global}. 
To address these issues, we implement deterministic APIs by ranking the matched items according to user ratings.

\paragraph{Complex User Goal.} To simulate more diverse user goals, we impose different relations for slot-value pairs. For example, in the restaurant searching scenarios, a user might want to eat Chinese food \texttt{(cuisine, equal\_to, Chinese)}, or do not want Chinese food \texttt{(cuisine, not, Chinese)}. Figure~\ref{fig:histo}.b shows the distribution of different relations in user goals.

\begin{table}[t]
\resizebox{\linewidth}{!}{
\begin{tabular}{@{}r|cccccccc@{}}
\toprule
\multicolumn{1}{l|}{} & \textbf{MultiWoZ} & \textbf{FRAMES} & \textbf{TM-1} & \textbf{SGD} & \textbf{STAR} & \textbf{RiSAWOZ} & \textbf{CrossWoz} & \textbf{BiToD} \\ \midrule
\textit{Language(s)} & EN & EN & EN & EN & EN & ZH & ZH & EN, ZH \\
\textit{Number of dialogues} & 8,438 & 1,369 & 13,215 & 16,142 & 5,820 & 10,000 & 5,012 & 5,787 \\
\textit{Number of domains} & 7 & 1 & 6 & 16 & 13 & 12 & 5 & 5 \\
\textit{Number of APIs} & 7 & 1 & 6 & 45 & 24 & 12 & 5 & 7 \\
\textit{Total number of turns} & 115,434 & 19,986 & 274,647 & 329,964 & 127,833 & 134,580 & 84,692 & 115,638 \\
\textit{Average turns / dialogues} & 13.46 & 14.6 & 21.99 & 20.44 & 21.71 & 13.5 & 16.9 & 19.98 \\
\textit{Slots} & 25 & 61 & - & 214 & - & 159 & 72 & 68* \\
\textit{Values} & 4,510 & 3,871 & - & 14,139 & - & 4,061 & 7,871 & 8,206* \\
\textit{Deterministic API} & \xmark & \xmark & \xmark & \xmark & \xmark & \xmark & \xmark & \cmark \\
\textit{Complex User Goal} & \xmark & \xmark & \xmark & \xmark & \cmark & \xmark & \xmark & \cmark \\
\textit{Mixed-Language Context} & \xmark & \xmark & \xmark & \xmark & \xmark & \xmark & \xmark & \cmark \\
\textit{Provided KB} & \cmark & \xmark & \xmark & \xmark & \cmark & \cmark & \cmark & \cmark \\ \bottomrule
\end{tabular}
}
\caption{Comparison of BiToD to previous ToD datasets. The numbers are provided for the \textit{\textbf{training set}} except for FRAMES and STAR. *We consider entities in different language as different slots and values.}
\label{tab:comparison}
\end{table}
\paragraph{Mixed-Language Context.} Our corpus contains code-switching utterances as some of the items in the knowledge base have mixed-language information. In the example in Figure~\ref{fig:example}.b, the system first recommends a restaurant called \textit{ChocoDuck Bistro} and the user asks for other options. Then the system searches other restaurants with an additional constraint \texttt{(restaurant\_name, not, ChocoDuck Bistro)}. In this example, both restaurants only have English names, which is a common phenomenon in multilingual regions like Hong Kong. Thus, ToD systems need to handle the mixed-language context to make correct API calls.

\paragraph{Cross-API Entity Carry-Over} Our corpus includes scenarios where the value of a slot is not presented in the conversation, and the system needs to carry over values from previous API results. In the example in Figure \ref{fig:example}.a, the user first finds and books a restaurant without specifying the location; then she (\includegraphics[height=0.29cm]{img/emoji/girl.png}) wants an attraction nearby the restaurant. In this case, the system needs to infer the attraction location (\texttt{\framebox[\width]{\textit{Wan Chai}}}) from the restaurant search result.

\section{Tasks \& Evaluations}
\subsection{Dialogue State Tracking}
Dialogue state tracking (DST), an essential task for ToD modelling, tracks the users' requirements over multi-turn conversations. DST labels provide sufficient information for a ToD system to issue APIs and carry out dialogue policies. In this work, we formulate a dialogue state as a set of slot-relation-value triples. 
We use Joint Goal Accuracy (\textbf{JGA}) to evaluate the performance of the DST. The model outputs are correct when all of the predicted slot-relation-value triples exactly match the oracle triples.

\subsection{End-to-End Task Completion}
A user's requests are fulfilled when the dialogue system makes correct API calls and correctly displays the requested information. We use the following automatic metrics to evaluate the performance of end-to-end task completion: 1) Task Success Rate (\textbf{TSR}): whether the system provides the correct entity and answers all the requested information of a given task, 2) Dialogue Success Rate (\textbf{DSR}): whether the system completes all the tasks in the dialogue, 3) API Call Accuracy ($\textbf{API}_{Acc}$): whether the system generates a correct API call, and 4) \textbf{BLEU}~\cite{papineni2002bleu}: measuring the fluency of the generated response.

\subsection{Evaluation Settings}
\paragraph{Monolingual.} Under the monolingual setting, models are trained and tested on either English or Chinese dialogues.
\paragraph{Bilingual.} Under the bilingual setting, models are trained on bilingual dialogues (full training set), and in the testing phase, the trained models are expected to handle dialogues in both languages simultaneously without any language identifiers.
\paragraph{Cross-lingual.} This setting simulates the condition of lacking data in a certain language, and we study how to transfer the knowledge from a high resource language to a low resource language. Models have full access to the source language in this setting but limited access to the target language (10\%).

\section{Proposed Baselines}
Our proposed baselines are based on the recent state-of-the-art end-to-end ToD modeling approach \textit{MinTL}~\cite{lin2020mintl} and cross-lingual transfer approach MTL~\cite{liu2020attention}. We report the hyper-parameters and training details in the Appendix~\ref{sec:infra}.

\begin{figure}[t]

  \begin{minipage}[c]{0.57\textwidth}
       \centering 
     \includegraphics[width=0.8\textwidth]{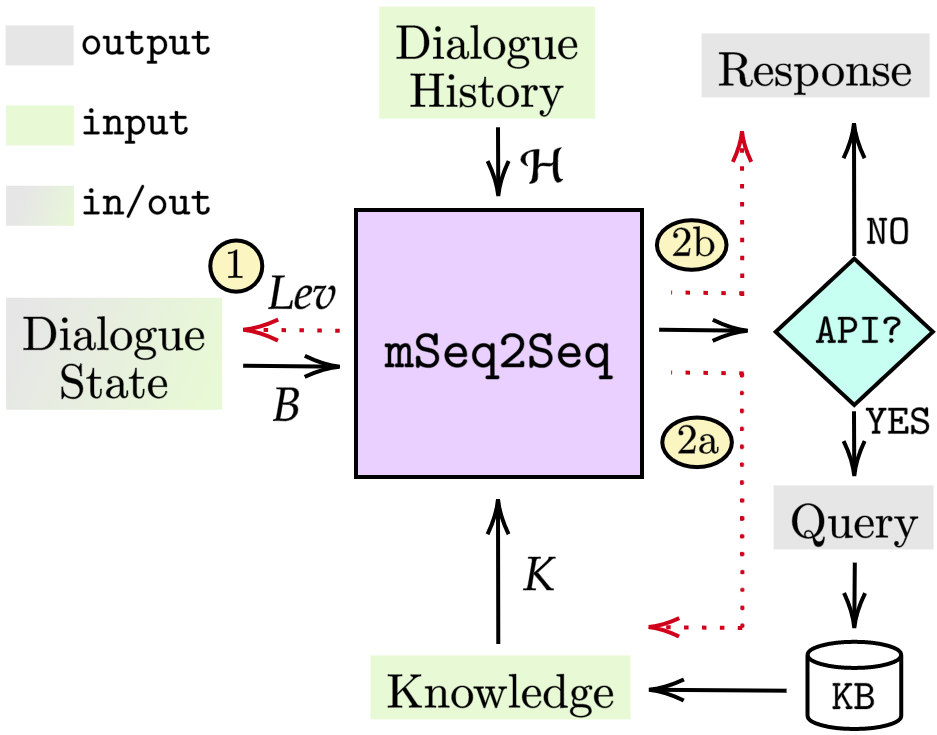}
  \end{minipage}
  \begin{minipage}[c]{0.4\textwidth}
    \caption{Model response generation workflow. Given the dialogue history $\mathcal{H}$, the knowledge $K$ (which can also be empty), and the dialogue state $B$, the $\mathrm{mSeq2Seq}$ 1) updates the dialogue state by generating $Lev$, and 2) generate a textual output and checks if it is an $\mathrm{API}$ or a $\mathrm{Response}$. If the output is an $\mathrm{API}$, (2a) the system queries the KB and updates the knowledge $K$; otherwise (2b) the $\mathrm{Response}$ is shown to the user. See this \href{https://github.com/HLTCHKUST/BiToD/blob/main/bitod.gif}{GIF} for more details.}
    \label{fig:model}
  \end{minipage}
\end{figure}
\paragraph{Notations.} 
We define a dialogue $\mathcal{D}=\{U_1,S_1, \dots, U_T, S_T\}$ as an alternating set of utterances from user and systems. At turn $t$, we denote a dialogue history as $\mathcal{H}_t=\{U_{t-w},S_{t-w}, \dots, S_{t-1},U_t\}$, where $w$ is the context window size. We denote the dialogue state and knowledge state at turn $t$ as $B_t$ and $K_t$, respectively.
\subsection{ToD Modeling}
Figure \ref{fig:model} describes the workflow of our baseline model. We initialize the dialogue state $B_0$ and knowledge state $K_0$ as empty strings. At turn $t$, the input of our model is the current dialogue history $H_t$, previous dialogue state $B_{t-1}$ and knowledge state $K_{t-1}$. Similar to the text-to-text transfer learning approach~\cite{raffel2020exploring}, we add a prompt $P_B = ``Track Dialogue State:"$ to indicate the generation task. Then, a multilingual sequence-to-sequence ($\mathrm{mSeq2Seq}$) model takes the flattened input sequence and outputs the \textit{Levenshtein Belief Spans} ($Lev_t$)~\cite{lin2020mintl}:
\begin{align*}
(1): \mathrm{Lev}_t = \mathrm{mSeq2Seq}(P_B, \mathcal{H}_t, B_{t-1}, K_{t-1}).
\end{align*}
The $Lev_t$ is a text span that contains the information for updating the dialogue state from $B_{t-1}$ to $B_t$.
The updated dialogue state $B_t$ and a response generation prompt, $P_R = "Response:"$, are used as input.
Then, the model will either generate an API name (2a) when an API call is needed at the current turn, or a plain text response directly returned to the user (2b). If the model generates an API name, it is
\begin{align*}
(2a): \mathrm{API} =\mathrm{mSeq2Seq}(P_R, \mathcal{H}_t, B_{t}, K_{t-1}),
\end{align*}
the system will query the API with the constraints in the dialogue state and update the knowledge state $K_{t-1} \rightarrow K_t$. The updated knowledge state and API name are incorporated into the model to generate the next turn response generation.
\begin{align*}
(2b): \mathrm{Response} = \mathrm{mSeq2Seq}(P_R, \mathcal{H}_t, B_{t}, K_{t}, \mathrm{API}).
\end{align*}
All the aforementioned generation process are based on a single $\mathrm{mSeq2Seq}$, and we initialized our model with two pre-trained models, \textbf{mT5}~\cite{raffel2020exploring} and \textbf{mBART}~\cite{liu2020multilingual}.

\subsection{Cross-lingual Transfer} 
Based on the modelling strategy mentioned above, we propose three baselines for the cross-lingual setting.
\paragraph{mSeq2seq.} Directly finetune the pre-trained mSeq2seq models like mBART and mT5 on the 10\% dialogue data in the target language.
\paragraph{Cross-lingual Pre-training (CPT).} First, pre-train the mBART and mT5 models on the source language, then finetune the models on the 10\% target language data.
\paragraph{Mixed-Language Pre-training (MLT).} To leverage the fact that our knowledge base contains the bilingual parallel information for most of the entities, we replace the entities in the source language data (both input sequence and output sequence) with their target language counterpart in our parallel knowledge base to generate the mixed-language training data. We first pre-train the mSeq2seq models with the generated mixed-language data, then finetune the models on the 10\% target language data.

\begin{table}[]
\resizebox{\linewidth}{!}{
\begin{tabular}{r|cccccccccc}
\toprule
\multicolumn{1}{l|}{} & \multicolumn{5}{c|}{\textit{English (EN)}} & \multicolumn{5}{c}{\textit{Chinese (ZH)}} \\ \midrule
\multicolumn{1}{l|}{\textit{\textbf{Models}}} & \textit{\textbf{TSR}} & \textit{\textbf{DSR}} & \textit{\textbf{API$_{\mathrm{Acc}}$}} & \textit{\textbf{BLEU}} & \multicolumn{1}{c|}{\textit{\textbf{JGA}}} & \textit{\textbf{TSR}} & \textit{\textbf{DSR}} & \textit{\textbf{API$_{\mathrm{Acc}}$}} & \textit{\textbf{BLEU}} & \textit{\textbf{JGA}} \\ \midrule 
 \multicolumn{11}{c}{\textit{\textbf{Monolingual}}} \\ \midrule 
\textit{mBART} & 56.00 & 33.71 & 57.03 & 35.34 & \multicolumn{1}{c|}{67.36} & 56.82 & 29.35 & 71.89 & 20.06 & \textbf{72.18} \\
\textit{mT5} & 69.13 & 47.51 & 67.92 & 38.48 & \multicolumn{1}{c|}{69.19} & 53.77 & 31.09 & 63.25 & 19.03 & 67.35 \\ \midrule 
\multicolumn{11}{c}{\textit{\textbf{Bilingual}}} \\ \midrule 
\textit{mBART} & 42.45 & 17.87 & 65.35 & 28.76 & \multicolumn{1}{c|}{69.37} & 40.39 & 16.96 & 65.37 & 5.23 & 69.50 \\
\textit{mT5} & \textbf{71.18} & \textbf{51.13} & \textbf{71.87} & \textbf{40.71} & \multicolumn{1}{c|}{\textbf{72.16}} & \textbf{57.24} & \textbf{34.78} & \textbf{65.54} & \textbf{22.45} & 68.70 \\ \midrule
\multicolumn{11}{c}{\textit{\textbf{Cross-lingual}}} \\ \midrule
\multicolumn{1}{l|}{} & \multicolumn{5}{c|}{ZH $\rightarrow$ EN (10\%)} & \multicolumn{5}{c}{EN $\rightarrow$ ZH (10\%)} \\ \midrule
\multicolumn{1}{l|}{\textit{\textbf{Models}}} & \textit{\textbf{TSR}} & \textit{\textbf{DSR}} & \textit{\textbf{API$_{\mathrm{Acc}}$}} & \textit{\textbf{BLEU}} & \multicolumn{1}{c|}{\textit{\textbf{JGA}}} & \textit{\textbf{TSR}} & \textit{\textbf{DSR}} & \textit{\textbf{API$_{\mathrm{Acc}}$}} & \textit{\textbf{BLEU}} & \textit{\textbf{JGA}} \\ \midrule
\textit{mBART} & 1.11 & 0.23 & 0.60 & 3.17 & \multicolumn{1}{c|}{4.64} & 0.00 & 0.00 & 0.00 & 0.01 & 2.14 \\
\textit{+ CPT} & 36.19 & 16.06 & 41.51 & 22.50 & \multicolumn{1}{c|}{42.84} & 24.64 & 11.96 & 29.04 & 8.29 & 28.57 \\
\multicolumn{1}{r|}{\textit{+ MLT}} & 33.62 & 11.99 & 41.08 & 20.01 & \multicolumn{1}{c|}{55.39} & 44.71 & 21.96 & \textbf{54.87} & 14.19 & \textbf{60.71} \\ \midrule 
\textit{mT5} & 6.78 & 1.36 & 17.75 & 10.35 & \multicolumn{1}{c|}{19.86} & 4.16 & 2.20 & 6.67 & 3.30 & 12.63 \\
\textit{+ CPT} & 44.94 & 24.66 & 47.60 & 29.53 & \multicolumn{1}{c|}{48.77} & 43.27 & 23.70 & 49.70 & 13.89 & 51.40 \\
\multicolumn{1}{r|}{\textit{+ MLT}} & \textbf{56.78} & \textbf{33.71} & \textbf{56.78} & \textbf{32.43} & \multicolumn{1}{c|}{\textbf{58.31}} & \textbf{49.20} & \textbf{27.17} & 50.55 & \textbf{14.44} & 55.05 \\ \bottomrule
\end{tabular}
}
\caption{Dialogue state tracking and end-to-end task completion results in monolingual, bilingual, and cross-lingual settings.}
\label{tab:main_results}
\end{table}

\section{Results \& Discussion}
The main results for DST and end-to-end task completion are reported in Table \ref{tab:main_results}. Note that the $API_{Acc}$ is highly correlated with the \textit{JGA} because the dialogue states contain constraints for issuing APIs. And, the \textit{DSR} is a more challenging metric compared to \textit{TSR} because the dialogue might contain 2-5 tasks.

\paragraph{Monolingual vs Bilingual.} Comparing the models that are trained under monolingual and bilingual setting, the latter can leverage more training data and handle tasks in both languages simultaneously without a language identifier. We observe that \textit{mT5} achieves better results in the bilingual settings, while \textit{mBART} performs better with monolingual training. The underlying reason might be the different pre-training strategies of the two mSeq2seq models. \textit{mBART} is pre-trained with language tokens in both the encoder and decoder, but in our bilingual setting, we do not provide any language information. Such a discrepancy does not exist in the \textit{mT5} model, as it is pre-trained without language tokens.

\paragraph{Cross-lingual.} We observe that it is difficult for the baseline models to converge with minimal training data (10\%) due to the complex ontology and diverse user goals. Interestingly, pre-training the mSeq2seq models on the source language improves both DST and task completion performance. Such results indicate the excellent cross-lingual transferability of multilingual language models. Furthermore, the mixed-language training strategy further improves the cross-lingual few shot performance, especially the \textit{JGA}, which suggests that the bilingual knowledge base can facilitate the cross-lingual knowledge transfer in the low resource scenario.

\paragraph{Limitations and Future Work.} The main limitation of this work is the low number of languages in the corpus due to the difficulty of collecting the knowledge base in languages other than English and Chinese in Hong Kong. in future work, we plan to extend the dataset to more languages including low resource languages in dialogue research (e.g., Indonesian), to better examine the cross-lingual transferability of end-to-end ToD systems. Another limitation is that the M2M data collection might not cover rare and unexpected user behaviours (e.g., non-collaborative dialogues), as dialogue simulators generate the dialogue outlines. However, we see {\dataset} as a necessary step for building robust multilingual ToD systems before tackling even more complex scenarios.

\section{Related Work}
Many datasets have been proposed in the past to support various assistant scenarios. In English, Wen et al. \cite{wen2016conditional} collected a single domain dataset with a Wizard-of-Oz (Woz) setup, which was latter extended to multi-domain by many follow-up works~\cite{budzianowski2018multiwoz,mosig2020star,moon2020situated,chen2021action}. Despite its effectiveness, Woz data collection method is expensive since two annotators need to be synchronized to conduct a conversation, and the other set of annotators need to annotate speech-act and dialogue states further. To reduce the annotation overhead (time and cost), Byrne et al. \cite{shah2018building} proposed a Machines Talking To Machines (M2M) self-chat annotations schema. Similarly, Rastongi et al.\cite{rastogi2020towards,byrne2019taskmaster} applied M2M to collect a large-scale schema-guided ToD dataset, and Kottur et al.\cite{kottur2021simmc} extended it to multimodal setting. In languages other than English, only a handful of datasets have been proposed. In Chinese, Zhu et al. ~\cite{zhu2020crosswoz}, and Quan et al.\cite{quan2020risawoz} proposed WoZ style datasets, and in German, the WMT 2020 Chat translated the dataset from Byrne et al. \cite{byrne2019taskmaster}. To the best of our knowledge, all the above-mentioned datasets are monolingual, thus making our BiToD dataset unique since it includes a bilingual setting and all the annotations needed for training an end-to-end task-oriented dialogue system. In the chit-chat setting, XPersona~\cite{lin2020xpersona} has a translation corpus in seven languages, but it is limited in the chit-chat domain. Finally, Razumovskaia et al.\cite{razumovskaia2021crossing} made an excellent summarization of the existing corpus for task-oriented dialogue systems, highlighting the need for multilingual benchmarks, like {\dataset}, for evaluating the cross-lingual transferability of end-to-end systems. 

\section{Conclusion}
We present {\dataset}, the first bilingual multi-domain dataset for end-to-end task-oriented dialogue modeling. {\dataset} contains over 7k multi-domain dialogues (144k utterances) with a large and realistic knowledge base. It serves as an effective benchmark for evaluating bilingual ToD systems and cross-lingual transfer learning approaches. We provide state-of-the-art baselines under three evaluation settings (monolingual, bilingual and cross-lingual). The analysis of our baselines in different settings highlights 1) the effectiveness of training a bilingual ToD system compared to two independent monolingual ToD systems, and 2) the potential of leveraging a bilingual knowledge base and cross-lingual transfer learning to improve the system performance under low resource conditions.


\bibliographystyle{unsrt} 
\bibliography{neurips_2021}

\section*{Checklist}


\begin{enumerate}

\item For all authors...
\begin{enumerate}
  \item Do the main claims made in the abstract and introduction accurately reflect the paper's contributions and scope?
    \answerYes{}
  \item Did you describe the limitations of your work?
    \answerYes{}
  \item Did you discuss any potential negative societal impacts of your work?
    \answerYes{} Section~\ref{sec:Ethics} for more information. 
  \item Have you read the ethics review guidelines and ensured that your paper conforms to them?
    \answerYes{}
\end{enumerate}

\item If you are including theoretical results...
\begin{enumerate}
  \item Did you state the full set of assumptions of all theoretical results?
    \answerNA{}
	\item Did you include complete proofs of all theoretical results?
    \answerNA{}
\end{enumerate}

\item If you ran experiments (e.g. for benchmarks)...
\begin{enumerate}
  \item Did you include the code, data, and instructions needed to reproduce the main experimental results (either in the supplemental material or as a URL)? 
    \answerYes{}
  \item Did you specify all the training details (e.g., data splits, hyperparameters, how they were chosen)?
    \answerYes{}
	\item Did you report error bars (e.g., with respect to the random seed after running experiments multiple times)?
    \answerNo{} 
	\item Did you include the total amount of compute and the type of resources used (e.g., type of GPUs, internal cluster, or cloud provider)?
    \answerYes{} Section~\ref{sec:infra} for more information. 
\end{enumerate}

\item If you are using existing assets (e.g., code, data, models) or curating/releasing new assets...
\begin{enumerate}
  \item If your work uses existing assets, did you cite the creators?
    \answerYes{}
  \item Did you mention the license of the assets?
    \answerYes{}
  \item Did you include any new assets either in the supplemental material or as a URL?
    \answerYes{}
  \item Did you discuss whether and how consent was obtained from people whose data you're using/curating?
    \answerYes{} 
  \item Did you discuss whether the data you are using/curating contains personally identifiable information or offensive content?
    \answerYes{}
\end{enumerate}

\item If you used crowdsourcing or conducted research with human subjects...
\begin{enumerate}
  \item Did you include the full text of instructions given to participants and screenshots, if applicable?
    \answerYes{}
  \item Did you describe any potential participant risks, with links to Institutional Review Board (IRB) approvals, if applicable?
    \answerNA{}
  \item Did you include the estimated hourly wage paid to participants and the total amount spent on participant compensation?
    \answerYes{}
\end{enumerate}

\end{enumerate}


\appendix

\section{Appendix}

\subsection{Ethics Statements} \label{sec:Ethics}

In this paper, we propose a new bilingual dataset for end-to-end task-oriented dialogue systems training and evaluation. Our dataset neither introduces any social/ethical, since we generate data with dialogue simulator and humanly paraphrase the utterances nor amplifies any bias. We do not foresee any direct social consequences or ethical issues. Furthermore, our proposed dataset encourages research in the cross-lingual few shot setting, where fewer data and resources are needed, rendering it energy-efficient models.
\subsection{Dataset documentation and intended uses}

We follow datasheets~\cite{gebru2018datasheets} for datasets guideline to document the following:
\subsubsection{Motivation}
\begin{itemize}[leftmargin=*]
    \item For what purpose was the dataset created? Was there a specific task in mind? Was there a specific gap that needed to be filled? 
    \begin{itemize}
        \item BiToD is created to benchmark the multilingual ability of end-to-end task oriented dialogue systems. Existing end-to-end benchmarks are limited to a single language (e.g., English or Chinese), thus BiToD fills the need of having a dataset for training and evaluating end-to-end task-oriented dialogue systems in the multilingual and cross-lingual settings. 
    \end{itemize}
    
    \item Who created the dataset (e.g., which team, research group) and on behalf of which entity (e.g., company, institution, organization)?
    \begin{itemize}
        \item HKUST CAiRE team and Alibaba team work together to create this dataset.
    \end{itemize}

    \item Who funded the creation of the dataset? If there is an associated grant, please provide the name of the grantor and the grant name and number.
    \begin{itemize}
        \item Alibaba team funded the creation of the dataset.
    \end{itemize}

\end{itemize}

\subsubsection{Composition}
\begin{itemize}[leftmargin=*]
    \item What do the instances that comprise the dataset represent (e.g., documents, photos, people, countries)? Are there multiple types of instances (e.g., movies, users, and ratings; people and interactions between them; nodes and edges)? Please provide a description.

    \begin{itemize}
        \item BiToD is made of conversations (text) between two speakers (user and assistant) and the textual knowledge in return from the API-call (tuple in a DB). BiToD also includes speech-acts for both user and systems, and dialogue state annotations.
    \end{itemize}

    \item How many instances are there in total (of each type, if appropriate)?
    \begin{itemize}
        \item BiToD has 7,232 dialogues with 144,798 utterances, in which 3,689 dialogues are in English and 3,543 dialogues are in Chinese. 
    \end{itemize}
    
    \item Does the dataset contain all possible instances or is it a sample (not necessarily random) of instances from a larger set? If the dataset is a sample, then what is the larger set? Is the sample representative of the larger set (e.g., geographic coverage)? If so, please describe how this representativeness was validated/verified. If it is not representative of the larger set, please describe why not (e.g., to cover a more diverse range of instances, because instances were withheld or unavailable).
    \begin{itemize}
        \item BiToD has been designed from scratch and thus contains all possible instances.  
    \end{itemize}

    \item What data does each instance consist of? “Raw” data (e.g., unprocessed text or images) or features? In either case, please provide a description.
    \begin{itemize}
        \item Each sample has raw text of conversations, speech-acts for both user and systems, dialogue state annotations, query, and knowledge bases return. 
    \end{itemize}

    \item Is there a label or target associated with each instance? If so, please provide a description.
    \begin{itemize}
        \item Each response is annotated with its speech-acts and the response it-self is target label. 
    \end{itemize}

    \item Is any information missing from individual instances? If so, please provide a description, explaining why this information is missing (e.g., because it was unavailable). This does not include intentionally removed information, but might include, e.g., redacted text.
    \begin{itemize}
        \item No, we included all the information we had.
    \end{itemize}

    \item Are relationships between individual instances made explicit (e.g., users’ movie ratings, social network links)? If so, please describe how these relationships are made explicit.
    \begin{itemize}
        \item No.
    \end{itemize}

    \item Are there recommended data splits (e.g., training, development/validation, testing)? If so, please provide a description of these splits, explaining the rationale behind them.
    \begin{itemize}
        \item Yes, we split the data into 80\% training, 8\% validation, and 12\% testing, resulting in 5,787 training dialogues, 542 validation dialogues, and 902 testing dialogues.
    \end{itemize}

    \item Are there any errors, sources of noise, or redundancies in the
dataset? If so, please provide a description.
    \begin{itemize}
        \item In 2.44\% of the dialogues, the annotators reported that the conversation did not sound formal enough, and in 1.11\% of the dialogues, the annotators reported that the dialogues are not valid -- did not sound coherent.
    \end{itemize}

    \item Is the dataset self-contained, or does it link to or otherwise rely on external resources (e.g., websites, tweets, other datasets)? If it links to or relies on external resources, a) are there guarantees that they will exist, and remain constant, over time; b) are there official archival versions of the complete dataset (i.e., including the external resources as they existed at the time the dataset was created); c) are there any restrictions] (e.g., licenses, fees) associated with any of the external resources that might apply to a future user? Please provide descriptions of all external resources and any restrictions associated with them, as well as links or other access points, as appropriate.
    \begin{itemize}
        \item Yes, BiToD is self-contained.
    \end{itemize}

    \item Does the dataset contain data that might be considered confidential (e.g., data that is protected by legal privilege or by doctorpatient confidentiality, data that includes the content of individuals’ non-public communications)? If so, please provide a description.
    \begin{itemize}
        \item No.
    \end{itemize}

    \item Does the dataset contain data that, if viewed directly, might be offensive, insulting, threatening, or might otherwise cause anxiety? If so, please describe why.
    \begin{itemize}
        \item No. 
    \end{itemize}

    \item Does the dataset relate to people? If not, you may skip the remaining questions in this section.
    \begin{itemize}
        \item No.
    \end{itemize}

    \item Does the dataset identify any subpopulations (e.g., by age, gender)? If so, please describe how these subpopulations are identified and provide a description of their respective distributions within the dataset.
    \begin{itemize}
        \item N/A
    \end{itemize}

    \item Is it possible to identify individuals (i.e., one or more natural persons), either directly or indirectly (i.e., in combination with other data) from the dataset? If so, please describe how.
    \begin{itemize}
        \item N/A
    \end{itemize}

    \item Does the dataset contain data that might be considered sensitive in any way (e.g., data that reveals racial or ethnic origins, sexual. orientations, religious beliefs, political opinions or union memberships, or locations; financial or health data; biometric or genetic data; forms of government identification, such as social security numbers; criminal history)? If so, please provide a description.
    \begin{itemize}
        \item N/A. 
    \end{itemize}
    

\end{itemize}

\subsubsection{Collection Process}
\begin{figure}
    \centering
    \includegraphics[width=\linewidth]{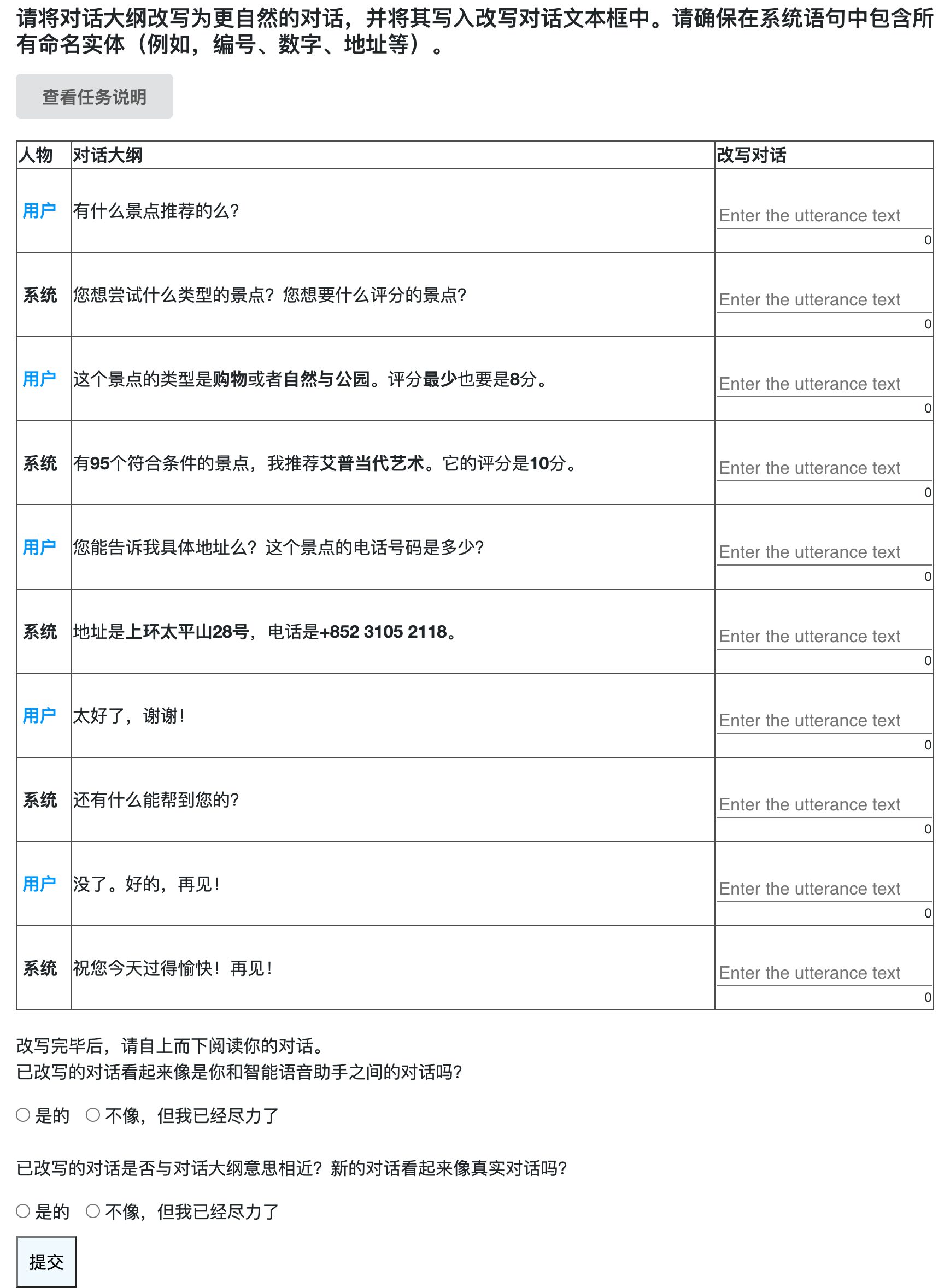}
    \caption{Interface of the Chinese dialogue paraphrasing. Crowd workers are asked to read the instructions before starting the task. All the system entities need to be reserved in the paraphrased dialogues.}
    \label{fig:zh_interface}
\end{figure}

\begin{itemize}[leftmargin=*]
    \item How was the data associated with each instance acquired? Was the data directly observable (e.g., raw text, movie ratings), reported by subjects (e.g., survey responses), or indirectly inferred/derived from other data (e.g., part-of-speech tags, model-based guesses for age or language)? If data was reported by subjects or indirectly inferred/derived from other data, was the data validated/verified? If so, please describe how.
    \begin{itemize}
        \item See main paper.
    \end{itemize}

    \item What mechanisms or procedures were used to collect the data (e.g., hardware apparatus or sensor, manual human curation, software program, software API)? How were these mechanisms or procedures validated? If the dataset is a sample from a larger set, what was the sampling strategy (e.g., deterministic, probabilistic with specific sampling probabilities)?
    \begin{itemize}
        \item Each utterance in the dialogue is paraphrased by Amazon Mechanical Turk for the English instances and AI-Speech~\footnote{http://www.aispeech.com/} for the Chinese instances. 
    \end{itemize}

    \item Who was involved in the data collection process (e.g., students, crowdworkers, contractors) and how were they compensated (e.g., how much were crowdworkers paid)?
    \begin{itemize}
        \item Crowdworkers. We paid them roughly \$10-12 per hour, calculated by the average time to write the paraphrase which is approximately 8 minutes.
    \end{itemize}

    \item Over what timeframe was the data collected? Does this timeframe match the creation timeframe of the data associated with the instances (e.g., recent crawl of old news articles)? If not, please describe the timeframe in which the data associated with the instances was created.
    \begin{itemize}
        \item The data was collected during February 2021 to May 2021.
    \end{itemize}

    \item Were any ethical review processes conducted (e.g., by an institutional review board)? If so, please provide a description of these review processes, including the outcomes, as well as a link or other access point to any supporting documentation.
    \begin{itemize}
        \item We have conducted an internal ethical review process by the HKUST ethical team.
    \end{itemize}

    \item Does the dataset relate to people? If not, you may skip the remainder of the questions in this section.
    \begin{itemize}
        \item No.
    \end{itemize}

    \item Did you collect the data from the individuals in question directly, or obtain it via third parties or other sources (e.g., websites)?
    \begin{itemize}
        \item N/A.
    \end{itemize}

    \item Were the individuals in question notified about the data collection? If so, please describe (or show with screenshots or other information) how notice was provided, and provide a link or other access point to, or otherwise reproduce, the exact language of the notification itself.
    \begin{itemize}
        \item Yes, the workers knew the data collection procedure. Screenshots are shown in Figure~\ref{fig:zh_interface}, Figure~\ref{fig:en_interface}, Figure~\ref{fig:en_instructions} and Figure~\ref{fig:en_annotation_example} in the Appendix.
    \end{itemize}

    \item Did the individuals in question consent to the collection and use of their data? If so, please describe (or show with screenshots or other information) how consent was requested and provided, and provide a link or other access point to, or otherwise reproduce, the exact language to which the individuals consented.
    \begin{itemize}
        \item AMT has its own data policy (\url{https://www.mturk.com/acceptable-use-policy}) and AI-Speech (\url{http://www.aispeech.com/}).
    \end{itemize}

    \item If consent was obtained, were the consenting individuals provided with a mechanism to revoke their consent in the future or for certain uses? If so, please provide a description, as well as a link or other access point to the mechanism (if appropriate).
    \begin{itemize}
        \item \url{https://www.mturk.com/acceptable-use-policy} and \url{http://www.aispeech.com/}.
    \end{itemize}

    \item Has an analysis of the potential impact of the dataset and its use on data subjects (e.g., a data protection impact analysis) been conducted? If so, please provide a description of this analysis, including the outcomes, as well as a link or other access point to any supporting documentation.
    \begin{itemize}
        \item N/A
    \end{itemize}

\end{itemize}

\subsubsection{Preprocessing/cleaning/labeling}
\begin{itemize}[leftmargin=*]
    \item Was any preprocessing/cleaning/labeling of the data done (e.g., discretization or bucketing, tokenization, part-of-speech tagging, SIFT feature extraction, removal of instances, processing of missing values)? If so, please provide a description. If not, you may skip the. remainder of the questions in this section. 
    \begin{itemize}
        \item No data cleaning or preprocessing is done for the released dataset since the dialogue data were generated by a simulator and only paraphrased by the workers.
    \end{itemize}
    
    \item Was the “raw” data saved in addition to the preprocessed/cleaned/labeled data (e.g., to support unanticipated future uses)? If so, please provide a link or other access point to the “raw” data.
    \begin{itemize}
        \item N/A.
    \end{itemize}

    \item Is the software used to preprocess/clean/label the instances available? If so, please provide a link or other access point.
    \begin{itemize}
        \item N/A.
    \end{itemize}

\end{itemize}

\subsubsection{Uses}
\begin{itemize}[leftmargin=*]
    \item Has the dataset been used for any tasks already? If so, please provide
a description.
    \begin{itemize}
        \item BiToD is a new dataset we collected for end-to-end task-oriented modeling and dialogue state tracking tasks. In this work, we build baseline models on BiToD for the mentioned tasks as a benchmark for future research.
    \end{itemize}

    \item Is there a repository that links to any or all papers or systems that use the dataset? If so, please provide a link or other access point.
    \begin{itemize}
        \item Yes, we release our dataset, code, and baseline models at \url{https://github.com/HLTCHKUST/BiToD}.
    \end{itemize}
    
    \item What (other) tasks could the dataset be used for?
    \begin{itemize}
        \item BiToD could be used for training dialogue policy by using the speech-act annotation, natural language generation modules, and user simulators. 
    \end{itemize}
    
    \item Is there anything about the composition of the dataset or the way it was collected and preprocessed/cleaned/labeled that might impact future uses? For example, is there anything that a future user might need to know to avoid uses that could result in unfair treatment of individuals or groups (e.g., stereotyping, quality of service issues) or other undesirable harms (e.g., financial harms, legal risks) If so, please provide a description. Is there anything a future user could do to mitigate these undesirable harms?
    \begin{itemize}
        \item No. 
    \end{itemize}
    
    \item Are there tasks for which the dataset should not be used? If so, please provide a description.
    \begin{itemize}
        \item No.
    \end{itemize}
    
\end{itemize}

\subsubsection{Distribution}
\begin{itemize}[leftmargin=*]
    \item Will the dataset be distributed to third parties outside of the entity (e.g., company, institution, organization) on behalf of which the dataset was created? If so, please provide a description.
    \begin{itemize}
        \item No.
    \end{itemize}
    
    \item How will the dataset will be distributed (e.g., tarball on website, API, GitHub)? Does the dataset have a digital object identifier (DOI)?
    \begin{itemize}
        \item It is released on Github at \url{https://github.com/HLTCHKUST/BiToD}. No DOI.
    \end{itemize}
    
    \item When will the dataset be distributed?
    \begin{itemize}
        \item It is released at our repository.
    \end{itemize}
    
    \item Will the dataset be distributed under a copyright or other intellectual property (IP) license, and/or under applicable terms of use (ToU)? If so, please describe this license and/or ToU, and provide a link or other access point to, or otherwise reproduce, any relevant licensing terms or ToU, as well as any fees associated with these restrictions.
    \begin{itemize}
        \item Apache License 2.0. \\
        \url{https://github.com/HLTCHKUST/BiToD/blob/main/LICENSE}
    \end{itemize}
    
    \item Have any third parties imposed IP-based or other restrictions on the data associated with the instances? If so, please describe these restrictions, and provide a link or other access point to, or otherwise reproduce, any relevant licensing terms, as well as any fees associated with these restrictions.
    \begin{itemize}
        \item No.
    \end{itemize}
    
    \item Do any export controls or other regulatory restrictions apply to the dataset or to individual instances? If so, please describe these restrictions, and provide a link or other access point to, or otherwise reproduce, any supporting documentation.
    \begin{itemize}
        \item No.
    \end{itemize}
    
\end{itemize}

\subsubsection{Maintenance}
\begin{itemize}[leftmargin=*]
    \item Who is supporting/hosting/maintaining the dataset?
    \begin{itemize}
        \item HKUST CAiRE research team. 
    \end{itemize}
    
    \item How can the owner/curator/manager of the dataset be contacted (e.g., email address)?
    \begin{itemize}
        \item Create an open issue on our Github repository or contact the authors (check author list email).
    \end{itemize}
    
    \item Is there an erratum? If so, please provide a link or other access point.
    \begin{itemize}
        \item No.
    \end{itemize}
    
    \item Will the dataset be updated (e.g., to correct labeling errors, add new instances, delete instances)? If so, please describe how often, by whom, and how updates will be communicated to users (e.g., mailing list, GitHub)?
    \begin{itemize}
        \item No. If we plan to update in the future, we will indicate the information on our Github repository.
    \end{itemize}
    
    \item If the dataset relates to people, are there applicable limits on the retention of the data associated with the instances (e.g., were individuals in question told that their data would be retained for a fixed period of time and then deleted)? If so, please describe these limits and explain how they will be enforced.
    \begin{itemize}
        \item No.
    \end{itemize}
    
    \item Will older versions of the dataset continue to be supported/hosted/maintained? If so, please describe how. If not, please describe how its obsolescence will be communicated to users.
    \begin{itemize}
        \item Yes. If we plan to update the data, we will keep the original version available and then release the follow-up version, for example, BiToD-2.0
    \end{itemize}
    
    \item If others want to extend/augment/build on/contribute to the dataset, is there a mechanism for them to do so? If so, please provide a description. Will these contributions be validated/verified? If so, please describe how. If not, why not? Is there a process for communicating/distributing these contributions to other users? If so, please provide a description.
    \begin{itemize}
        \item Yes, they can submit a Github pull request or contact us privately. 
    \end{itemize}
    
\end{itemize}

\subsection{Accessibility}

\begin{enumerate}[leftmargin=*]
  \item Links to access the dataset and its metadata. \\
  \url{https://github.com/HLTCHKUST/BiToD}
  
  \item The data is saved in a json format, where an example is shown in the README.md file. 
  
  \item HKUST CAiRE team will maintain this dataset on the official company Github account.
  
  \item Apache License 2.0. \\
 \url{https://github.com/HLTCHKUST/BiToD/blob/main/LICENSE}

\end{enumerate}

\subsection{Data Usage}

The authors bear all responsibility in case of violation of rights.

\subsection{Training Details}\label{sec:infra}
We implement our baselines based on the huggingface Transformers~\cite{wolf2019huggingface}. In all of our experiments, we set the dialogue context window size $w=2$ and we use the pre-trained model mT5-small and mBART-large. They are trained with batch size 128 using an AdamW~\cite{loshchilov2018decoupled} optimizer with the initial learning rate of $0.0005$ and $0.0001$ respectively. In monolingual and bilingual settings, all the models are trained for 8 epochs, while in cross-lingual setting, the models are first trained on source language dialogues for 8 epochs and then fine tune the model on target language for 10 epochs. We use 2 NVIDIA V100 GPUs for mBART training and 2 1080Ti for mT5 training. All the trainings take less than 10 hours. We use greedy decoding in test time. More training information is available in \url{https://github.com/HLTCHKUST/BiToD}.

\begin{table}[]
\begin{tabular}{@{}r|ccc@{}}
\toprule
\textbf{Domain} & \multicolumn{1}{c|}{\textbf{API}} & \multicolumn{1}{c|}{\textbf{Informable Slots}} & \textbf{Requestable Slots} \\ \midrule
\multirow{3}{*}{\textit{Restaurant}} & \multicolumn{1}{c|}{search} & \multicolumn{1}{c|}{\begin{tabular}[c]{@{}c@{}}dietary\_restrictions, \\ cuisine, name, \\ price\_level, \\ location, rating\end{tabular}} & \begin{tabular}[c]{@{}c@{}}cuisine, name, \\ phone\_number, \\ available\_options, \\ address, rating\end{tabular} \\ \cmidrule(l){2-4} 
 & \multicolumn{1}{c|}{booking} & \multicolumn{1}{c|}{\begin{tabular}[c]{@{}c@{}}name, date, user\_name, \\ time, number\_of\_people,\end{tabular}} & ref\_number \\ \midrule
\textit{Attraction} & \multicolumn{1}{c|}{search} & \multicolumn{1}{c|}{type, name, location, rating} & \begin{tabular}[c]{@{}c@{}}name, phone\_number, \\ available\_options,\\ address, rating\end{tabular} \\ \midrule
\textit{Metro} & \multicolumn{1}{c|}{MTR} & \multicolumn{1}{c|}{departure, destination} & \begin{tabular}[c]{@{}c@{}}shortest\_path, \\ estimated\_time, price\end{tabular} \\ \midrule
\textit{Weather} & \multicolumn{1}{c|}{search} & \multicolumn{1}{c|}{day, city} & \begin{tabular}[c]{@{}c@{}}day, max\_temp, \\ weather, city, \\ min\_temp\end{tabular} \\ \midrule
\multirow{4}{*}{\textit{Hotel}} & \multicolumn{1}{c|}{search} & \multicolumn{1}{c|}{\begin{tabular}[c]{@{}c@{}}name, stars, price\_level, \\ location, rating\end{tabular}} & \begin{tabular}[c]{@{}c@{}}available\_options, name, \\ price\_per\_night, rating\end{tabular} \\ \cmidrule(l){2-4} 
 & \multicolumn{1}{c|}{booking} & \multicolumn{1}{c|}{\begin{tabular}[c]{@{}c@{}}name, number\_of\_rooms, \\ start\_day, user\_name, \\ start\_month, \\ number\_of\_nights\end{tabular}} & ref\_number \\ \midrule \midrule
Act type & \multicolumn{3}{l}{\begin{tabular}[c]{@{}l@{}}affirm, offer, request\_more, notify\_fail, confirm, inform\_intent, goodbye, negate,\\ inform, request, notify\_success, thank\_you, greeting, request\_update\end{tabular}} \\ \midrule
Rel. type & \multicolumn{3}{l}{equal\_to, not, less\_than, at\_least, one\_of} \\ \bottomrule
\end{tabular}
\caption{BiToD English ontology. }
\label{tab:ontology}
\end{table}

\begin{figure}
    \centering
    \includegraphics[width=\linewidth]{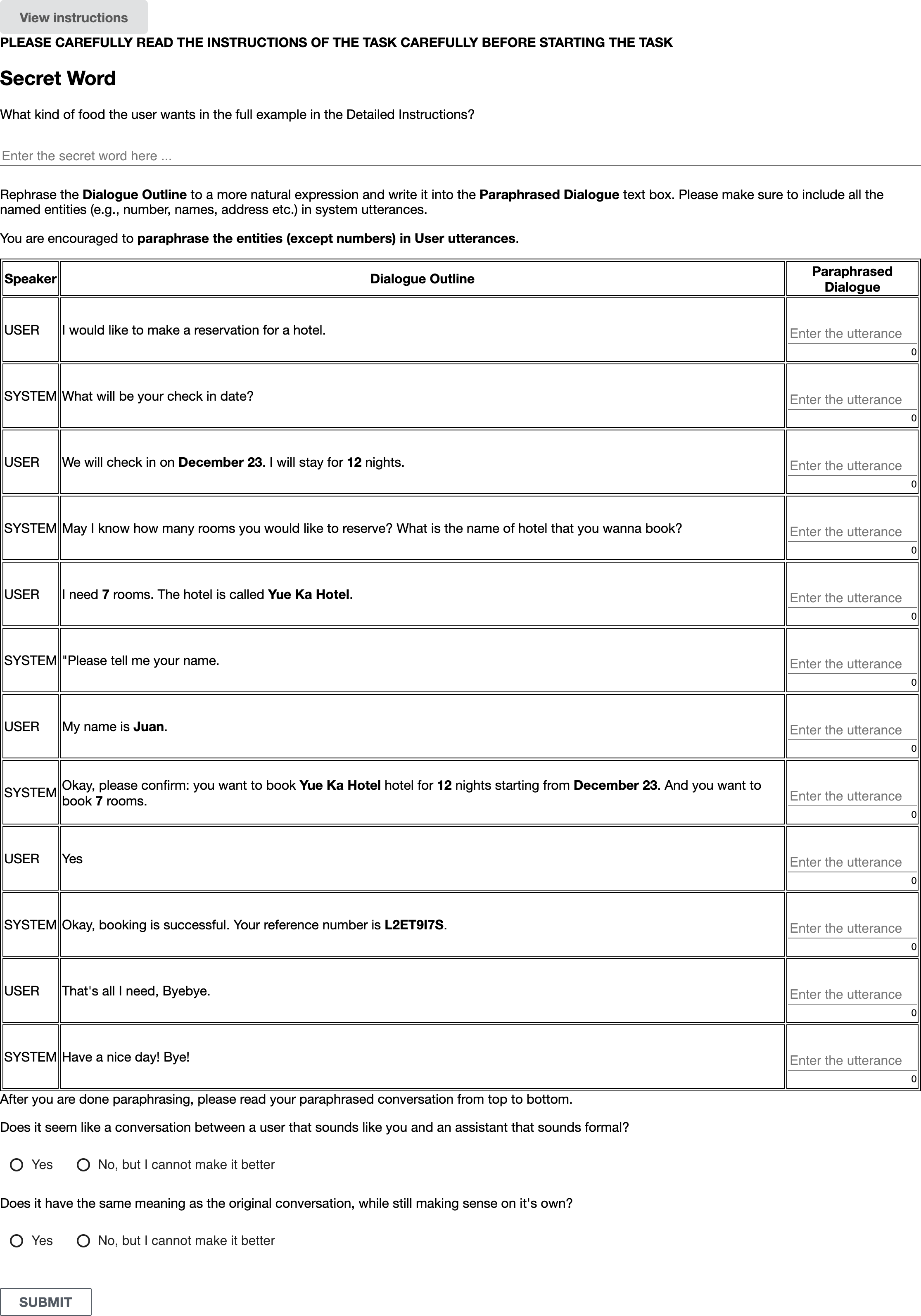}
    \caption{Interface of the English dialogue paraphrasing. Crowd workers are asked to read the instructions before starting the task. All the system entities need to be reserved in the paraphrased dialogues}
    \label{fig:en_interface}
\end{figure}

\begin{figure}
    \centering
    \includegraphics[width=\linewidth]{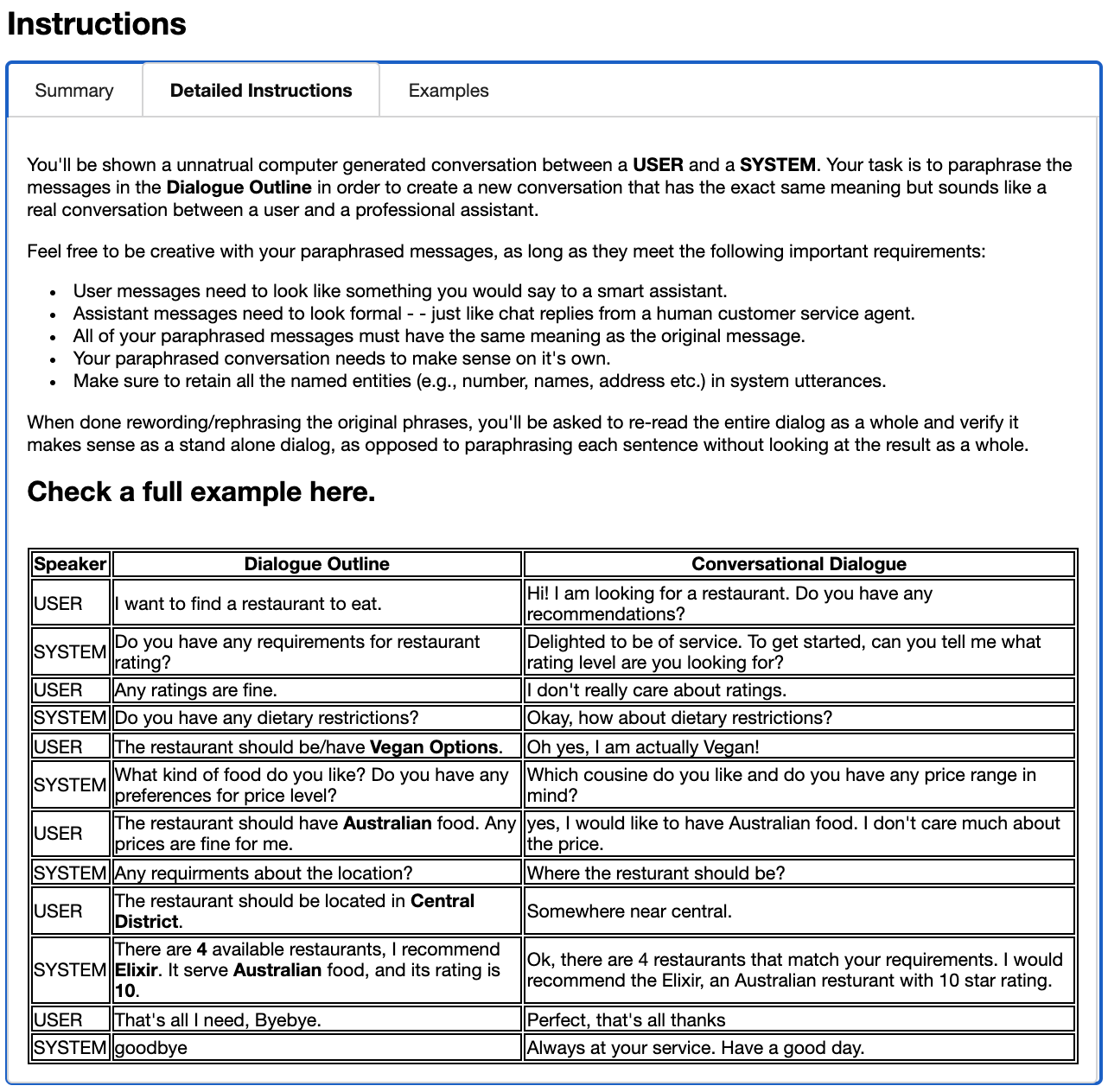}
    \caption{Detailed instructions for English dialogue paraphrasing.}
    \label{fig:en_instructions}
\end{figure}

\begin{figure}
    \centering
    \includegraphics[width=\linewidth]{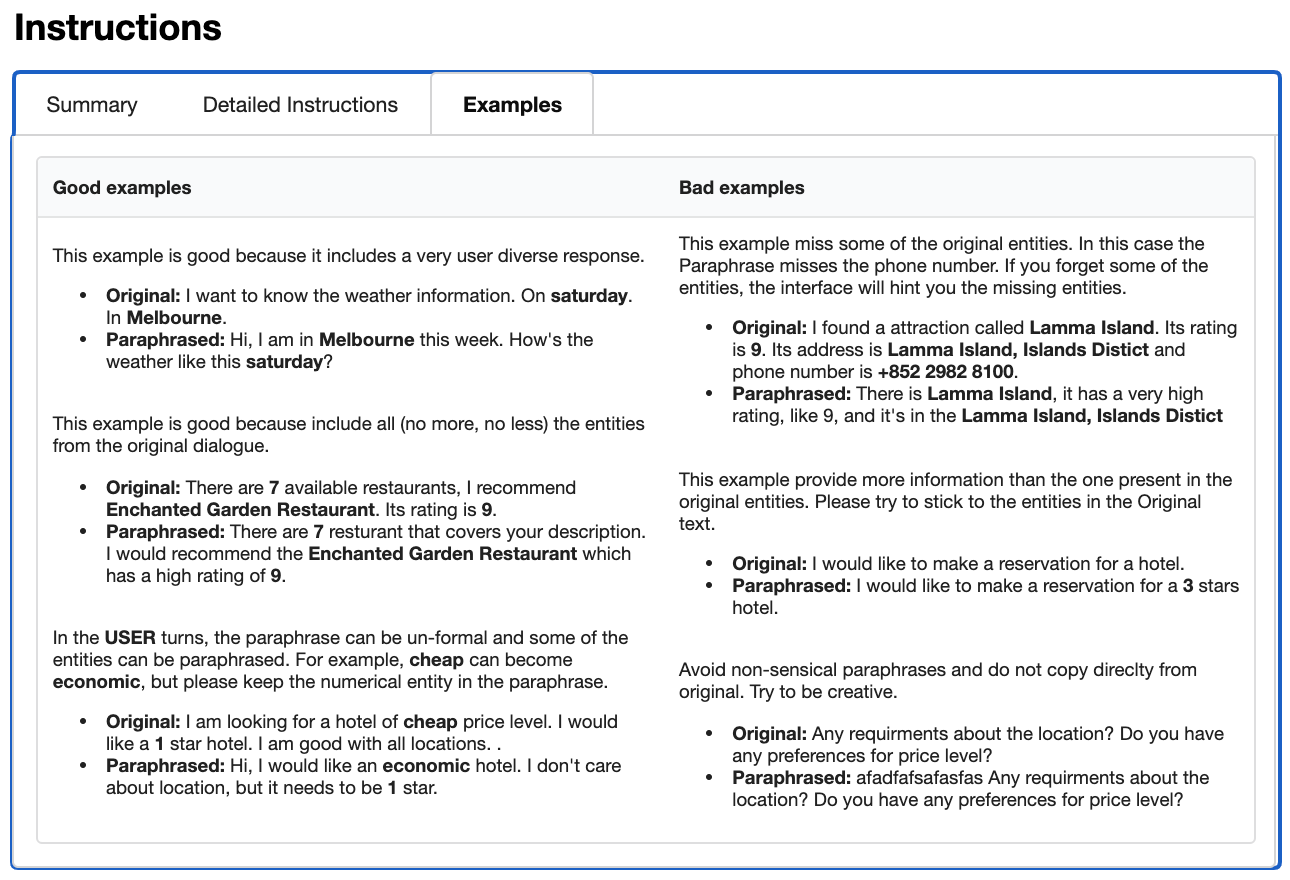}
    \caption{Examples for English dialogue paraphrasing.}
    \label{fig:en_annotation_example}
\end{figure}

\end{document}